\documentclass[10pt]{article}

\usepackage[utf8]{inputenc}
\usepackage[T1]{fontenc}
\usepackage{lmodern}
\usepackage{graphicx}
\usepackage{hyperref}
\usepackage{booktabs}
\usepackage{amsmath}
\usepackage{natbib}
\usepackage[margin=1in]{geometry}
\usepackage{float}

\setcitestyle{authoryear,round,citesep={;},aysep={,},yysep={;}}

\title{DepthCharge: A Domain-Agnostic Framework for Measuring Depth-Dependent Knowledge in Large Language Models}

\author{Alexander Sheppert \\
Legacy Health \\ Capitol Technology University \\
\texttt{asheppert@captechu.edu}}

\date{}

\begin{document}

\maketitle

\begin{abstract}
Large Language Models appear competent when answering general questions but often fail when pushed into domain-specific details. No existing methodology provides an out-of-the-box solution for measuring how deeply LLMs can sustain accurate responses under adaptive follow-up questioning across arbitrary domains.

We present DepthCharge, a domain-agnostic framework that measures knowledge depth through three innovations: adaptive probing that generates follow-up questions based on concepts the model actually mentions, on-demand fact verification from authoritative sources, and survival statistics with constant sample sizes at every depth level. The framework can be deployed on any knowledge domain with publicly verifiable facts, without requiring pre-constructed test sets or domain-specific expertise. DepthCharge results are relative to the evaluator model used for answer checking, making the framework a tool for comparative evaluation rather than absolute accuracy certification.

Empirical validation across four diverse domains (Medicine, Constitutional Law, Ancient Rome, and Quantum Computing) with five frontier models demonstrates that DepthCharge reveals depth-dependent performance variation hidden by standard benchmarks. Expected Valid Depth (EVD) ranges from 3.45 to 7.55 across model-domain combinations, and model rankings vary substantially by domain, with no single model dominating all areas. Cost-performance analysis further reveals that expensive models do not always achieve deeper knowledge, suggesting that domain-specific evaluation is more informative than aggregate benchmarks for model selection in professional applications.
\end{abstract}

\section{Introduction}

\subsection{Motivation}

Large language models are increasingly deployed in high-stakes professional settings: physicians use them to interpret clinical findings, lawyers to research case precedents, and researchers to explore literature. These applications require not just surface-level fluency but deep, accurate knowledge that holds up under scrutiny. A model that correctly answers ``What causes influenza?'' but fails on follow-up questions about antiviral medications presents a serious risk. For example, when asked ``You mentioned influenza viruses. What mutations allow the virus to resist oseltamivir (Tamiflu)?'' a model may falter despite its confident initial response. Users may trust the model based on its initial competence, unaware that reliability degrades as questions become more specific.

No existing methodology provides an out-of-the-box solution for comparing the depth of understanding across LLMs in arbitrary domains. Current benchmarks like MMLU, TruthfulQA, and MedQA test broad coverage across many topics but rarely probe deeply within any single domain \citep{chen2025llmbenchmarksurvey}. More critically, they require extensive manual curation: subject matter experts must craft questions, verify answers, and validate difficulty levels before any testing can begin. This creates a significant barrier when organizations need to evaluate LLM knowledge in specialized fields. Recent surveys have catalogued hundreds of such benchmarks \citep{wu2025sebenchmarks}, yet the fundamental problem of depth evaluation remains unaddressed.

Recent work has begun addressing the problem of static benchmarks. LiveBench \citep{livebench2024} generates fresh questions monthly to prevent test set contamination. OKBench \citep{okbench2024} creates questions from daily news to ensure novelty. These approaches address the critical problem that models can be trained directly on benchmark content, artificially inflating scores without genuine capability improvement \citep{yang2025contamination, wu2025contamination, xu2025pitfalls}. However, these dynamic benchmarks still test breadth rather than depth: they generate novel questions across many topics rather than drilling progressively deeper into specific domains.

Consider the practical problem: a hospital wants to evaluate which LLM best understands cardiology for clinical decision support. Recent medical LLM benchmarks have found average performance of only 57\% across clinical domains, with safety performance lagging effectiveness \citep{chen2025medicalbenchmark}. A law firm wants to assess LLM knowledge of tax law for research assistance. A pharmaceutical company wants to measure LLM understanding of drug interactions. Each evaluation currently requires either building a custom benchmark from scratch (a process that can take months and substantial expert resources) or accepting that existing benchmarks test surface-level knowledge that may not predict performance on specialized follow-up questions \citep{li2025medcheck, wang2025constructvalidity}.

DepthCharge addresses this gap by providing a unified, domain-agnostic testing framework. Given any knowledge domain, the system automatically generates questions of increasing difficulty, verifies answers against authoritative sources, and produces standardized metrics for cross-model comparison. The same methodology that evaluates cardiology knowledge can evaluate tax law, pharmacology, or any other domain where verifiable facts exist.

\subsection{The Surface Competence Illusion}

Static benchmarks create what we term a ``surface competence illusion'': models score well on varied shallow questions while harboring significant gaps in specialized knowledge that only emerge through follow-up questioning. Because benchmark questions are pre-scripted, they cannot adapt to probe the specific concepts a model claims to understand. A model might correctly state that ``influenza is caused by orthomyxoviruses'' while being unable to explain the role of neuraminidase in viral release when asked.

We ask a specific question: how reliably can models handle adaptive follow-up drilling into topics they claim expertise on? By ``drilling,'' we mean the process of extracting concepts from the model's answer and asking deeper questions about those specific concepts. By ``adaptive,'' we mean that the path of questioning depends on what the model actually says, not a fixed script. The term ``depth charge'' reflects how the framework distributes questions across surviving knowledge paths, expanding its probing at each depth level like an underwater depth charge detonating at a target depth.

\subsection{Framework Overview}

We present DepthCharge, a testing framework that combines three key innovations:

First, adaptive drilling: rather than asking pre-scripted questions, we extract concepts from the model's answers and drill into whatever the model mentions. If the model emphasizes ``respiratory transmission,'' we drill into respiratory transmission. If it emphasizes ``viral structure,'' we drill into viral structure. This mimics how an expert would probe a student's knowledge.

Second, verified ground truth: before asking any question, we search for verifiable facts about the topic. Every question has a known correct answer, eliminating the need for post-hoc judgment about correctness. We search Wikipedia for common knowledge, and use specialized search APIs for professional-level facts.

Third, survival statistics with constant sample size: at each depth, we ask exactly N questions (default 30), distributed across surviving paths. If 10 paths survive, each gets 3 questions. If 3 paths survive, each gets 10 questions. This maintains statistical power at every depth, preventing single-path variance from dominating results.

\subsection{Contributions}

Our contributions are:

First, we introduce DepthCharge, a domain-agnostic adaptive drilling system that can evaluate LLM knowledge depth in any subject area without requiring pre-constructed test sets.

Second, we apply survival analysis methods with proper sample sizes. Rather than reporting accuracy on paths that may dwindle to single cases, we ensure 30 questions at every depth, yielding statistically meaningful survival rates and tighter confidence intervals.

Third, we map depth to difficulty through specificity tiers with configurable passes per tier: COMMON, TEXTBOOK, PROFESSIONAL, SPECIALIST, and CUTTING\_EDGE. Each tier can have Q passes (default Q=3), allowing fine-grained measurement of knowledge depth within each specificity level.

Fourth, we validate the framework empirically across five commercially available frontier models (the leading large language models available via API at time of testing) and four diverse domains, demonstrating that DepthCharge reveals depth-dependent performance variation hidden by aggregate accuracy metrics and that model rankings vary substantially by domain.

\section{Related Work}

\subsection{LLM Benchmarks and Evaluation}

The evaluation of large language models has evolved substantially in recent years. Chen et al. \citep{chen2025llmbenchmarksurvey} provide a comprehensive survey of 283 representative benchmarks, categorizing them into general capabilities, domain-specific, and target-specific evaluations. They identify critical problems including inflated scores from data contamination, cultural and linguistic biases, and lack of process credibility evaluation. Wu et al. \citep{wu2025sebenchmarks} review 291 benchmarks specifically for software engineering tasks.

Static benchmarks face the fundamental problem of test set contamination, where models may have been trained on benchmark content \citep{yang2025contamination, wu2025contamination}. Xu et al. \citep{xu2025pitfalls} demonstrate that open benchmarks create substantial risks of data leakage. Zhang et al. \citep{zhang2025arxivroll} propose ArxivRoll, a dynamic evaluation framework that generates new benchmarks every six months using recent arXiv articles.

To address contamination, dynamic benchmarks have emerged. LiveBench \citep{livebench2024} generates fresh questions monthly from recent information sources, achieving ICLR 2025 Spotlight recognition. OKBench \citep{okbench2024} provides fully automated, on-demand benchmark generation that can create new evaluations daily. These approaches ensure novelty but primarily test breadth rather than depth.

\subsection{Cognitive and Taxonomic Evaluation}

Recent work has applied Bloom's taxonomy to LLM evaluation with revealing results. Liu et al. \citep{liu2025bloomtaxonomy} at COLING 2025 systematically analyzed how existing benchmarks cover cognitive skill levels. They evaluated multiple LLMs across 12 benchmarks mapped to Bloom's six cognitive levels, finding that LLMs generally perform better on lower taxonomy levels (Remember, Understand) and that higher levels (Metacognitive, Create, Evaluate) remain largely uncovered by existing benchmarks. Their analysis revealed a consistent performance degradation as cognitive complexity increases.

Chen et al. \citep{chen2025bloomwise} developed BloomWise, a cognitively-inspired prompting technique that progresses through cognitive operations from basic to advanced, demonstrating that structured cognitive scaffolding improves LLM performance on complex reasoning tasks.

Zhang et al. \citep{zhang2025bloomapr} introduced BloomAPR, a taxonomy-based framework for automated program repair that assesses LLM capabilities across progressively complex reasoning levels. Their experiments on standard benchmarks revealed a striking pattern: LLMs effectively memorize patterns at the Remember level (81\% accuracy) but struggle significantly when asked to apply knowledge with minor variations (43\% at Apply level). This 38-percentage-point drop between memorization and application suggests that high performance on factual recall does not predict ability to use that knowledge flexibly.

Wang et al. \citep{wang2025multicognitive} introduced a multi-cognitive-level evaluation framework specifically for medical domains based on the revised Bloom's taxonomy \citep{krathwohl2001revision}, finding similar patterns of strong recall but weaker application in clinical reasoning tasks.

\subsection{Multi-Turn and Adaptive Evaluation}

Multi-turn dialogue evaluation has received increasing attention. Chen et al. \citep{chen2025multiturn} survey recent advances in multi-turn LLM interactions, noting that MT-Bench is limited to two-turn dialogues. Wu et al. \citep{wu2025lostinconversation} demonstrate that LLMs make early assumptions, prematurely propose solutions, and have difficulty course-correcting when provided with new information. Liu et al. \citep{liu2025dialogueevaluator} propose efficient dialogue evaluators that aggregate multiple LLM judge feedback.

Adaptive testing approaches from psychometrics have been applied to LLMs. Wang et al. \citep{wang2025atlas} introduce ATLAS, an adaptive testing framework based on Item Response Theory (IRT) that reduces required evaluation items by up to 90\% while maintaining measurement precision. Liu et al. \citep{liu2025adaptivedistraction} propose adaptive distraction generation using tree search to stress-test contextual robustness, finding average performance drops of over 45\% across mainstream models.

IRT-based methods have gained traction for LLM evaluation. Chen et al. \citep{chen2025lart} propose Latency-Response Theory (LaRT) that jointly models response accuracy and chain-of-thought length. Wang et al. \citep{wang2025irtnet} introduce IrtNet, learning compact representations of LLM abilities using a Mixture-of-Experts architecture. Li et al. \citep{li2025psnirt} propose PSN-IRT for enhanced IRT frameworks, revealing significant shortcomings in measurement quality across 11 benchmarks. Zhang et al. \citep{zhang2025irtrouter} develop IRT-Router for interpretable multi-LLM routing.

\subsection{Medical and Domain-Specific Benchmarks}

Medical LLM evaluation has advanced substantially. Wang et al. \citep{wang2025llmevalmed} introduce LLMEval-Med with 2,996 questions from real-world electronic health records. Li et al. \citep{li2025medcheck} present MedCheck, a lifecycle-oriented framework with 46 medically-tailored criteria, finding widespread issues including disconnect from clinical practice and data integrity crises. Chen et al. \citep{chen2025medicalbenchmark} report that average LLM performance across clinical domains was 57.2\%, with safety performance (54.7\%) lower than effectiveness (62.3\%). Zhang et al. \citep{zhang2025medkgeval} introduce MedKGEval, a knowledge graph-driven multi-agent framework for realistic medical dialogues. Wang et al. \citep{wang2025constructvalidity} argue that benchmarks should prioritize construct validity through empirical evaluation.

Professional forum benchmarks have also emerged. Wu et al. \citep{wu2025lpfqa} present LPFQA, a benchmark derived from authentic professional forum discussions covering 7 domains with 430 curated tasks, exposing performance gaps on tasks requiring deep domain reasoning.

\subsection{Hallucination and Factuality}

Hallucination detection remains a central challenge. Chen et al. \citep{chen2025hallulens} introduce HalluLens at ACL 2025, providing a comprehensive hallucination benchmark with dynamic test set generation. Wang et al. \citep{wang2025hallucinationsurvey} provide a comprehensive survey introducing structured taxonomies of detection and mitigation approaches. Liu et al. \citep{liu2025unifiedhallucination} demonstrate that hybrid approaches integrating hallucination detection and fact verification achieve state-of-the-art performance.

Knowledge graph integration improves factuality. Zhang et al. \citep{zhang2025peek} propose PEEK for efficient knowledge probing, achieving up to 90\% accuracy in predicting LLM knowledge. Li et al. \citep{li2025alignedllm} introduce ALIGNed-LLM for improving factuality through knowledge graph alignment. Wang et al. \citep{wang2025kgfactuality} demonstrate that inference-time knowledge graph construction improves factual accuracy. Chen et al. \citep{chen2025llmkg} survey the synthesis of LLMs and knowledge graphs for question answering.

\subsection{Question Generation}

Automatic question generation supports scalable evaluation. Liu et al. \citep{liu2025aqag} implement automatic question-answer generation using fine-tuned generative LLMs. Wang et al. \citep{wang2025qaevaluation} demonstrate that LLM-generated assessments can match human-authored tests in psychometric performance. Zhang et al. \citep{zhang2025generatethenvalidate} propose generate-then-validate approaches using small language models with LLM validation. Li et al. \citep{li2025contextqa} combine consistency verification with constraint checking for high-quality test case generation.

\subsection{Positioning of DepthCharge}

DepthCharge differs from existing approaches in several ways. Unlike static benchmarks, it generates questions adaptively based on model responses. Unlike dynamic benchmarks like LiveBench and OKBench that test breadth, DepthCharge tests depth through progressive drilling. Unlike IRT-based approaches that estimate ability from existing item pools, DepthCharge generates questions on-demand for any domain. Unlike multi-turn dialogue evaluations that assess conversational coherence, DepthCharge specifically measures knowledge reliability under sustained factual questioning. The combination of adaptive drilling, on-demand fact verification, and cumulative survival statistics provides a unique framework for domain-agnostic depth evaluation.

\section{Methodology}

\subsection{Domain Requirements}

DepthCharge can be applied to any knowledge domain that satisfies the following requirements:

First, the domain must have publicly accessible factual content at the common knowledge level. Currently, this means Wikipedia coverage of the topic. Wikipedia provides the verified ground truth for COMMON tier (depths 1-3) and TEXTBOOK tier (depths 4-6) questions. Topics without Wikipedia articles cannot be tested at these foundational levels.

Second, for testing beyond textbook level, the domain should have academic or professional literature. PROFESSIONAL tier (depths 7-9) questions draw from clinical guidelines, professional standards, and authoritative references. SPECIALIST tier (depths 10-12) requires peer-reviewed literature. CUTTING\_EDGE tier (depths 13+) requires recent publications from the past two years. Domains without such literature will reach their testing ceiling at the TEXTBOOK tier.

Third, the domain must contain verifiable factual claims. Subjective domains (aesthetic preferences, philosophical opinions) or domains where ground truth is contested cannot be reliably tested. The framework works best for scientific, technical, legal, and medical domains where expert consensus establishes verifiable facts.

These requirements are not onerous: most professional and academic domains satisfy all three. Medical specialties, legal areas, scientific fields, engineering disciplines, and technical subjects all have Wikipedia coverage, professional literature, and verifiable facts. The framework deliberately targets the domains where LLM accuracy matters most: high-stakes professional applications.

\subsection{Adaptive Drilling Process}

DepthCharge operates through iterative adaptive drilling:

\begin{enumerate}
    \item At depth 1, generate diverse initial branches (e.g., ``Influenza causes,'' ``Influenza symptoms,'' ``Influenza types'')
    \item For each branch, search for a verifiable fact from appropriate sources
    \item Generate a question from the verified fact
    \item Quiz the target model
    \item Score the answer against the verified fact
    \item For correct answers, extract concepts mentioned and prepare to drill into them
    \item At the next depth, distribute N questions across surviving paths
    \item Continue until survival rate drops below threshold (default 20\%)
\end{enumerate}

The key innovation is that drilling directions are determined by what the model says. If the model mentions ``neuraminidase inhibitors'' in its answer about treatment, we drill into neuraminidase inhibitors specifically. This creates personalized probing paths that reveal each model's actual knowledge boundaries.

\subsection{On-Demand Fact Verification}

Rather than pre-building a knowledge graph, we search for facts in real-time as we drill. This on-demand approach enables domain-agnostic testing: the same framework works for cardiology, tax law, or quantum physics without modification.

For COMMON tier (depths 1-3): Wikipedia summary API provides basic facts that any educated person would know. Wikipedia has known limitations: coverage varies by topic, recent events may be incomplete, and vandalism occasionally introduces errors. However, for established factual content at the foundational level, Wikipedia offers the best combination of coverage, accessibility, and editorial oversight available. Studies have found Wikipedia's accuracy comparable to traditional encyclopedias for scientific topics, and its rapid correction of errors provides reasonable reliability for common knowledge facts. For organizations requiring higher accuracy guarantees, the framework supports substituting curated knowledge bases.

For TEXTBOOK tier (depths 4-6): Wikipedia detailed sections and full article content provide student-level knowledge, including technical details, historical context, and established mechanisms. The same limitations apply, though detailed sections undergo less frequent editing and may contain more stable content.

For PROFESSIONAL and higher tiers, we use retrieval-augmented systems that search the web in real-time for each query and return information grounded in retrieved sources. This is not circular reasoning (LLM verifying LLM): the retrieval system synthesizes information from web documents, academic papers, and authoritative sources, providing citation-backed responses. The grounding in retrieved documents means the facts come from the underlying source material, not from the model's training data.

For PROFESSIONAL tier (depths 7-9): The search is guided to prioritize authoritative sources such as medical associations, regulatory bodies, and professional organizations.

For SPECIALIST tier (depths 10-12): Searches target peer-reviewed literature, molecular mechanisms, and expert-level details. Search queries are structured to find information from academic databases and specialized publications.

For CUTTING\_EDGE tier (depths 13+): Searches focus on recent publications, emphasizing papers and findings from the past two years. This tests whether models have knowledge of the latest developments in their training data.

If we cannot find a verifiable fact for a drilling direction, that probe is marked as failed (no ground truth available). This is rare for common topics but increases at specialist levels, which itself provides information about the boundaries of publicly verifiable knowledge in a domain.

\subsection{Question Distribution}

To maintain statistical power, we always ask exactly N questions per depth (default 30). This distribution mechanism gives the framework its name: like a naval depth charge that detonates and expands at a specific depth, DepthCharge distributes its probing across available knowledge paths at each level, expanding coverage where paths converge and maintaining breadth where they diverge.

\begin{itemize}
    \item If 30+ paths survive: sample 30 paths, ask 1 question each
    \item If 10 paths survive: ask 3 questions per path
    \item If 5 paths survive: ask 6 questions per path
    \item If 1 path survives: ask 30 questions about different aspects
\end{itemize}

When a single path needs multiple questions, we generate multiple drilling directions from that path's answer. This ensures that accuracy at depth D is always based on 30 observations, not 1. The expansion at each depth prevents statistical fragility: without this mechanism, a single surviving path's single correct answer could show misleading ``100\% accuracy'' at deep levels despite minimal statistical power.

\subsection{Difficulty Mapping}

Each depth maps to a specificity tier that determines how specialized the knowledge must be. The framework uses a configurable number of passes per tier (Q), allowing fine-grained measurement within each specificity level before advancing to more specialized content.

The specificity tiers represent increasingly specialized knowledge:

\begin{itemize}
    \item COMMON: General public knowledge (Wikipedia summaries)
    \item TEXTBOOK: University student level (Wikipedia detailed sections)
    \item PROFESSIONAL: Practitioner level (clinical guidelines, professional standards)
    \item SPECIALIST: Expert level (peer-reviewed literature, specialist references)
    \item CUTTING\_EDGE: Researcher level (recent publications from the past two years)
\end{itemize}

With Q passes per tier, the depth-to-tier mapping is:

\begin{table}[h]
\centering
\begin{tabular}{lll}
\toprule
Depth & Tier & Pass \\
\midrule
1 & COMMON & P1 \\
2 & COMMON & P2 \\
3 & COMMON & P3 \\
4 & TEXTBOOK & P1 \\
5 & TEXTBOOK & P2 \\
6 & TEXTBOOK & P3 \\
7 & PROFESSIONAL & P1 \\
8 & PROFESSIONAL & P2 \\
9 & PROFESSIONAL & P3 \\
10+ & SPECIALIST+ & ... \\
\bottomrule
\end{tabular}
\caption{Mapping from depth to difficulty level with Q=3 passes per tier.}
\end{table}

This creates a monotonic difficulty progression where deeper drilling requires progressively more specialized knowledge. Multiple passes within each tier allow thorough exploration of knowledge at that specificity level before advancing. All questions are factual in nature, asking ``What is...'', ``Which...'', or ``Name the...'' to ensure verifiable correct answers from authoritative sources.

\subsection{Question Difficulty Normalization}

Because drilling paths are adaptive (determined by each model's responses), different models may receive different specific questions at the same depth. This raises a concern about comparability: one model's depth-5 question might be inherently easier than another's.

We mitigate this through difficulty-constrained question generation. The question generator receives explicit instructions for the target specificity tier:

\begin{itemize}
    \item Tier constraints: Questions must draw from tier-appropriate sources. COMMON questions use Wikipedia summaries; TEXTBOOK questions use detailed Wikipedia sections; PROFESSIONAL questions require practitioner-level sources.
    \item Factual focus: All questions are factual in nature, asking ``What is...'', ``Which...'', or ``Name the...'' to ensure verifiable correct answers from authoritative sources.
    \item Fact verification: Every question is generated from a pre-verified fact at the appropriate tier, ensuring the difficulty reflects the knowledge source rather than arbitrary question framing.
\end{itemize}

While this does not guarantee identical questions across models, it ensures that all questions at a given depth target the same difficulty specification. Crucially, evaluations are fully reproducible: given the same random seed, the same model probed on the same topic receives identical questions, enabling direct replication and longitudinal comparison. The variation in specific topics across models (neuraminidase vs hemagglutinin, for example) reflects each model's actual knowledge boundaries rather than arbitrary path divergence. Future work could explore fixed question sets for direct comparability, though this would sacrifice the adaptive drilling that reveals model-specific knowledge gaps.

\subsection{Survival Metrics}

Let $n_d = N$ (constant, e.g., 30) be questions asked at depth $d$, and $c_d$ be correct answers. The accuracy at depth $d$ is:

\begin{equation}
    A(d) = \frac{c_d}{n_d}
\end{equation}

We chose cumulative (multiplicative) survival over alternatives such as average accuracy or windowed survival for several reasons. First, it models the realistic scenario where a user asks a series of follow-up questions: each incorrect answer terminates that line of inquiry, and the user's trust degrades based on the entire interaction history, not just the most recent response. Second, it prevents masking of early errors by later successes: a model that fails 50\% of initial questions cannot compensate by performing well on the surviving half. Third, it creates clearer separation between models: small accuracy differences compound across depths, making EVD more sensitive to reliability differences than simple accuracy averaging. Alternative formulations (windowed survival over the last K depths, weighted survival that discounts early errors) may be appropriate for use cases where recent performance matters more than history.

We compute cumulative survival $S(d)$ as the product of accuracies at all depths up to and including $d$:

\begin{equation}
    S(d) = \prod_{i=1}^{d} A(i) = S(d-1) \times A(d)
\end{equation}

This cumulative formulation reflects that incorrect answers terminate their paths: a model that achieves 80\% accuracy at depth 1 and 50\% accuracy at depth 2 has cumulative survival of $0.8 \times 0.5 = 0.4$ (40\%), not 50\%. The failed paths from depth 1 cannot produce correct answers at depth 2, so the maximum possible survival at depth 2 is bounded by survival at depth 1.

Expected Valid Depth (EVD) is the area under the cumulative survival curve:

\begin{equation}
    \text{EVD} = \sum_{d=1}^{D} S(d)
\end{equation}

where $D$ is the maximum depth reached before cumulative survival drops below threshold.

We stop drilling when $S(d) < \theta$ (default $\theta = 0.20$), indicating cumulative survival has degraded below useful levels.

\begin{figure}[H]
\centering
\includegraphics[width=\textwidth]{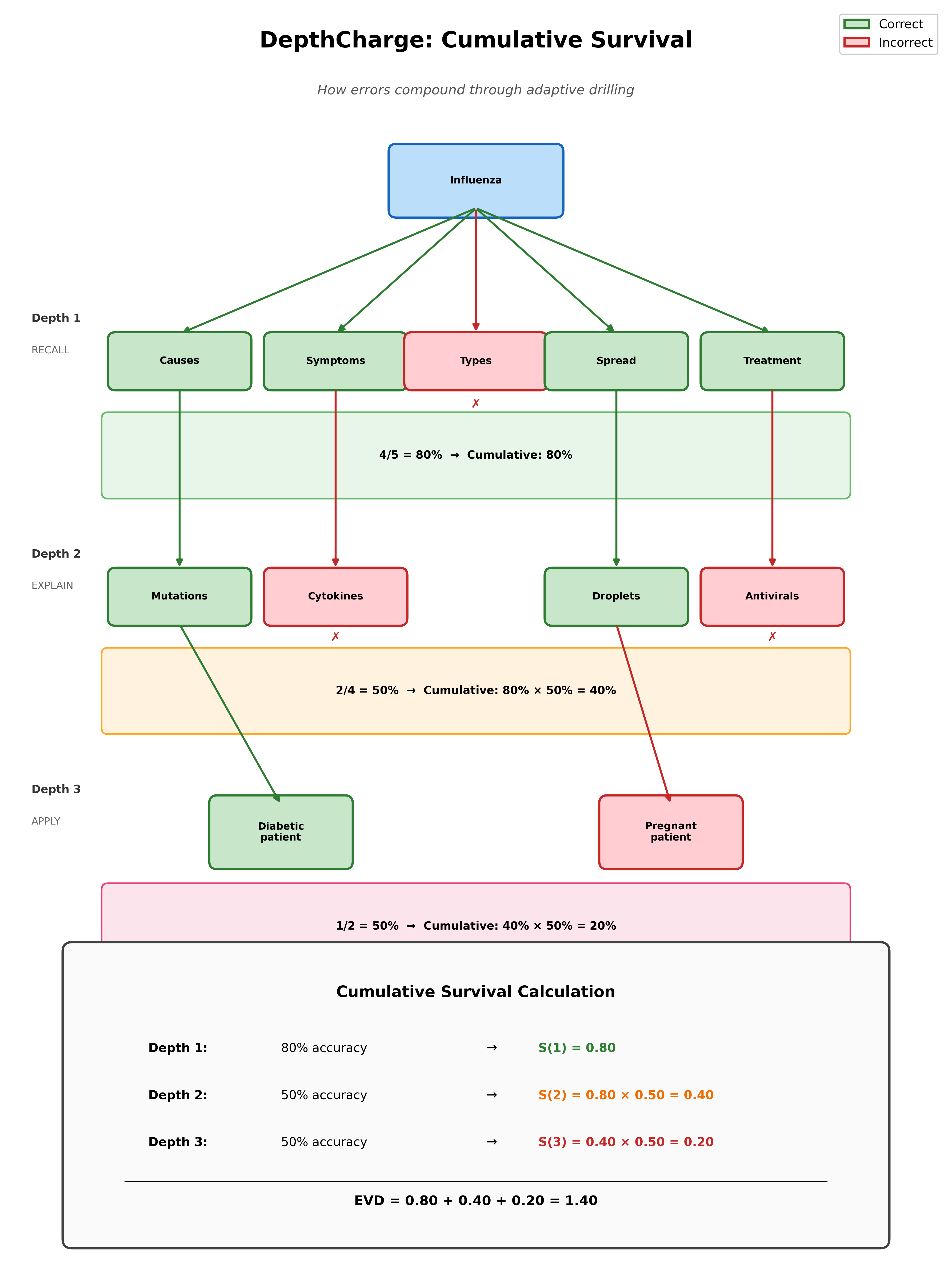}
\caption{Tree diagram showing adaptive drilling with cumulative survival calculation. Green nodes indicate correct answers (paths continue), red nodes indicate incorrect answers (paths terminate). Cumulative survival at each depth equals the product of all accuracies up to that point, reflecting that errors permanently reduce survival probability.}
\label{fig:tree}
\end{figure}

Figure \ref{fig:tree} illustrates this process with an example. At depth 1, 4 of 5 questions are answered correctly (80\% accuracy), yielding 80\% cumulative survival. At depth 2, only the 4 correct paths continue; 2 of 4 questions are correct (50\% accuracy), yielding $0.8 \times 0.5 = 0.40$ (40\%) cumulative survival. At depth 3, 1 of 2 remaining paths is correct (50\% accuracy), and cumulative survival drops to $0.40 \times 0.5 = 0.20$ (20\%). The EVD for this example is $0.80 + 0.40 + 0.20 = 1.40$.

\begin{figure}[H]
\centering
\includegraphics[width=\textwidth]{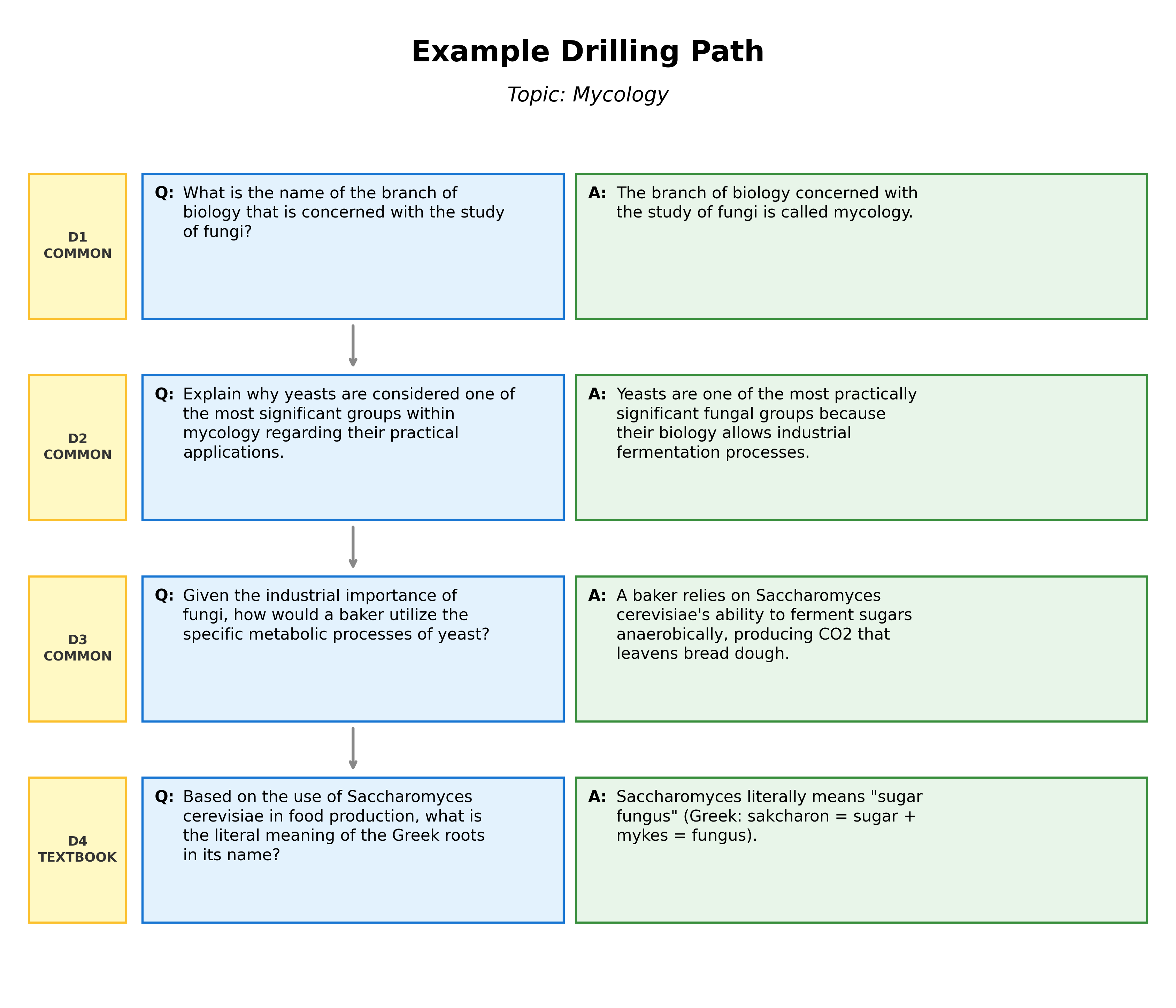}
\caption{Example drilling path showing question-answer pairs at increasing depths. D1-D3 (COMMON tier) ask increasingly specific questions about foundational concepts. D4-D6 (TEXTBOOK tier) probe detailed knowledge requiring deeper familiarity with the subject. Each question builds on concepts from the previous answer.}
\label{fig:example_path}
\end{figure}

\section{Experimental Setup}

\subsection{Models}

We test five commercially available frontier models (the leading large language models available via API at time of testing) via a unified API, anonymized as Models A through E. These models span different providers and price points, ranging from \$0.09 to \$5.80 per domain evaluation (Table \ref{tab:costs}). A separate model serves as the evaluator for concept extraction, question generation, and answer scoring.

\subsection{Protocol}

Each model is tested with the following parameters:
\begin{itemize}
    \item 30 questions per depth level (N=30)
    \item 3 passes per specificity tier (Q=3)
    \item Maximum depth 15 (allowing progression through CUTTING\_EDGE tier)
    \item Cumulative survival threshold 20\%
    \item Random seed 42 for reproducibility
\end{itemize}

Testing stops when cumulative survival falls below the threshold, reflecting accumulated errors across all depths. This cumulative approach means models must maintain consistent accuracy throughout the drilling process. The larger sample size (N=30) provides tighter confidence intervals compared to smaller samples, with 95\% CI of approximately $\pm$12 percentage points at 50\% accuracy.

\subsection{Computational Cost}

Each model-domain evaluation requires approximately 1,000 API calls across the target model, extractor model, and fact verification service. A typical evaluation reaching 7-11 depths completes in approximately 60-90 minutes. Table \ref{tab:costs} shows the relative costs per model for a single domain evaluation.

\begin{table}[h]
\centering
\begin{tabular}{lrr}
\toprule
Model & Cost/Domain & Relative Cost \\
\midrule
Model D & \$0.09 & 1.0x (baseline) \\
Model E & \$0.61 & 6.8x \\
Model A & \$0.96 & 10.7x \\
Model C & \$4.34 & 48.2x \\
Model B & \$5.80 & 64.4x \\
\bottomrule
\end{tabular}
\caption{Relative costs per model-domain evaluation with N=30 questions per depth.}
\label{tab:costs}
\end{table}

Cost scales primarily with the target model's per-token pricing. Model D provides exceptional value at under \$0.10 per domain while achieving competitive depth performance. The full 20-evaluation suite (5 models $\times$ 4 domains) costs approximately \$47 USD total and completes in approximately 90 minutes of wall-clock time when run in parallel.

\subsection{Domains Tested}

To demonstrate domain-agnostic applicability, we evaluate across four diverse domains:
\begin{itemize}
    \item Medicine (Medical Science): Clinical and biomedical knowledge spanning diagnosis, treatment, and pathophysiology
    \item Constitutional Law (Legal): Foundational legal domain with interpretive complexity
    \item Ancient Rome (History): Classical history with extensive primary and secondary sources
    \item Quantum Computing (Physics/Computer Science): Technical domain combining abstract physics and computing
\end{itemize}

These four domains were selected to demonstrate the framework's versatility across fundamentally different knowledge types. The selection spans medical science, law, history, and physics/computing, representing distinct epistemological traditions: empirical science (Medicine), interpretive legal reasoning (Constitutional Law), historical scholarship (Ancient Rome), and theoretical physics combined with engineering (Quantum Computing). This diversity demonstrates that DepthCharge can evaluate any domain meeting the requirements in Section 3.1, enabling organizations to deploy the framework on their specific domains of interest without modification.

\section{Results}

The following results validate DepthCharge as a framework by demonstrating that it (1) produces consistent, reproducible measurements across diverse domains, (2) reveals depth-dependent performance variation hidden by aggregate accuracy metrics, and (3) shows that model rankings are domain-dependent, motivating domain-specific rather than aggregate evaluation.

\subsection{Framework Validation: Cross-Domain Results}

\begin{table}[h]
\centering
\small
\begin{tabular}{llrrrrrr}
\toprule
Model & Domain & EVD & Max D & Acc. & COMMON & TEXT & PROF \\
\midrule
A & Medicine & 7.55 & 11 & 87\% & 100\% & 76\% & 77\% \\
D & Constitutional Law & 6.94 & 11 & 82\% & 97\% & 80\% & 49\% \\
A & Constitutional Law & 6.49 & 11 & 82\% & 97\% & 84\% & 49\% \\
D & Quantum Computing & 6.21 & 9 & 83\% & 100\% & 71\% & 62\% \\
C & Quantum Computing & 6.06 & 9 & 82\% & 97\% & 79\% & 46\% \\
B & Ancient Rome & 5.36 & 8 & 84\% & 100\% & 68\% & 40\% \\
A & Quantum Computing & 5.33 & 8 & 84\% & 100\% & 67\% & 63\% \\
D & Ancient Rome & 5.24 & 8 & 82\% & 99\% & 67\% & 47\% \\
C & Constitutional Law & 5.22 & 10 & 78\% & 96\% & 66\% & 50\% \\
C & Medicine & 5.17 & 9 & 84\% & 96\% & 78\% & 54\% \\
B & Quantum Computing & 5.06 & 8 & 81\% & 96\% & 67\% & 53\% \\
B & Medicine & 5.41 & 9 & 83\% & 97\% & 81\% & 38\% \\
A & Ancient Rome & 4.96 & 7 & 85\% & 100\% & 67\% & 47\% \\
D & Medicine & 4.85 & 8 & 81\% & 99\% & 77\% & 20\% \\
C & Ancient Rome & 4.79 & 8 & 81\% & 97\% & 68\% & 38\% \\
B & Constitutional Law & 4.16 & 7 & 80\% & 96\% & 71\% & 17\% \\
E & Ancient Rome & 4.07 & 7 & 80\% & 97\% & 66\% & 30\% \\
E & Constitutional Law & 3.62 & 6 & 78\% & 97\% & 52\% & -- \\
E & Medicine & 3.52 & 6 & 75\% & 97\% & 71\% & 20\% \\
E & Quantum Computing & 3.45 & 7 & 73\% & 92\% & 56\% & 27\% \\
\bottomrule
\end{tabular}
\caption{DepthCharge results across all 20 model-domain combinations sorted by EVD. EVD = Expected Valid Depth. Max D = maximum depth reached. TEXT = TEXTBOOK, PROF = PROFESSIONAL. With N=30 questions per depth, 95\% confidence intervals are approximately $\pm$12 percentage points at 50\% accuracy (Wilson score intervals).}
\label{tab:results}
\end{table}

\begin{figure}[H]
\centering
\includegraphics[width=\textwidth]{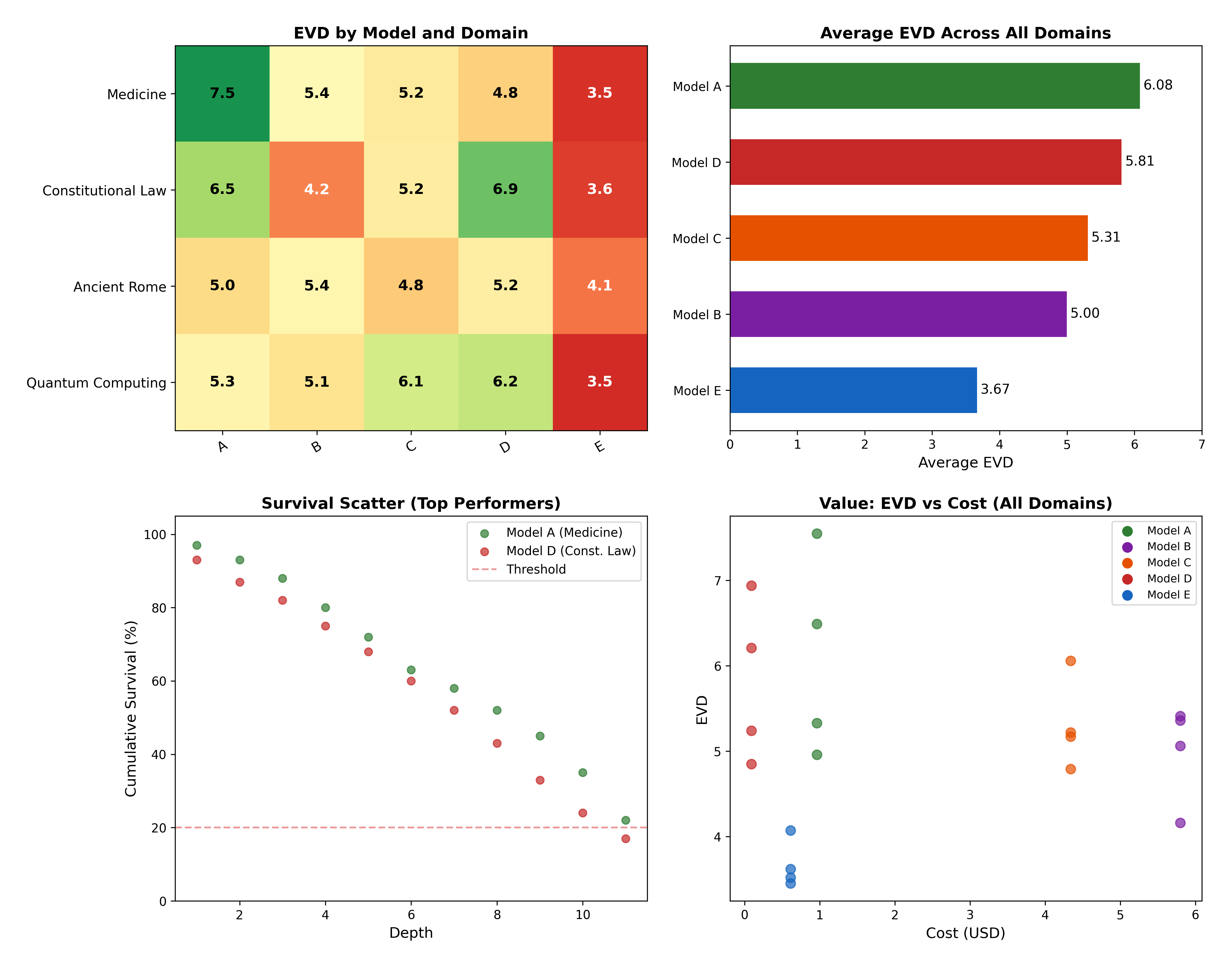}
\caption{Comprehensive comparison across all model-domain combinations. Top left: EVD heatmap showing performance variation. Top right: Average EVD rankings. Bottom left: Survival scatter comparing top performers. Bottom right: Value analysis (EVD vs cost).}
\label{fig:comparison}
\end{figure}

Table~\ref{tab:results} presents results across all 20 model-domain combinations. The framework produces a wide range of EVD values (3.45 to 7.55), demonstrating sufficient sensitivity to differentiate model capabilities across domains and depth levels.

Two patterns validate the framework's utility. First, EVD reveals variation that aggregate accuracy obscures: overall accuracy ranges narrowly (73--87\%) across model-domain combinations, while EVD varies by more than a factor of two (3.45 to 7.55). The cumulative survival formulation amplifies small per-depth accuracy differences into meaningful depth distinctions. Second, model rankings are domain-dependent: no single model achieves the highest EVD across all four domains. Model A achieves the highest EVD on Medicine (7.55) but not on Constitutional Law, where Model D achieves the highest (6.94). Model B achieves the highest EVD on Ancient Rome (5.36) but ranks lower on other domains. This domain-dependent variation confirms that depth-based evaluation captures information absent from aggregate benchmarks.

The framework also reveals cost-performance tradeoffs invisible in standard evaluations. Model D achieves the second-highest mean EVD (5.81) at approximately 10x lower cost per evaluation than Model A (mean EVD 6.08), suggesting that model selection for professional applications benefits from domain-specific depth evaluation rather than reliance on aggregate rankings.

All models perform well on COMMON tier (92--100\%), with accuracy degrading progressively through TEXTBOOK (52--84\%) and PROFESSIONAL (17--77\%) tiers, confirming that the tier-based difficulty progression produces monotonically increasing challenge as intended.

\subsection{Tier-Based Difficulty Validation}

The tier progression (COMMON, TEXTBOOK, PROFESSIONAL) is designed to produce monotonically increasing difficulty. The results confirm this design: accuracy degrades consistently across tiers for all models and domains.

COMMON (depths 1--3): All models achieve high accuracy (92--100\%), confirming that the COMMON tier appropriately targets foundational knowledge. The narrow spread across models at this tier (4--8 percentage points) indicates that current frontier models reliably handle general knowledge, consistent with existing benchmark findings.

TEXTBOOK (depths 4--6): Accuracy drops to 52--84\%, and the spread across models widens substantially. This tier represents the critical transition where the framework begins to differentiate models. The wider spread confirms that TEXTBOOK-level questions probe beyond what all models handle uniformly.

PROFESSIONAL (depths 7--9): All five models reach this tier on at least some domains, with accuracy ranging from 17\% to 77\%. The wide range demonstrates that the framework successfully identifies model-specific knowledge ceilings. Notably, a model's PROFESSIONAL accuracy varies substantially across domains (e.g., Model D achieves 62\% on Quantum Computing but 20\% on Medicine), confirming that depth profiles are domain-specific.

\subsection{Statistical Reliability}

With N=30 questions per depth, we report 95\% Wilson score confidence intervals for accuracy estimates. Table \ref{tab:ci} shows representative confidence intervals at different accuracy levels.

\begin{table}[h]
\centering
\begin{tabular}{lrr}
\toprule
Observed Accuracy & 95\% CI Lower & 95\% CI Upper \\
\midrule
50\% (15/30) & 33\% & 67\% \\
70\% (21/30) & 52\% & 83\% \\
80\% (24/30) & 62\% & 91\% \\
90\% (27/30) & 74\% & 97\% \\
100\% (30/30) & 88\% & 100\% \\
\bottomrule
\end{tabular}
\caption{95\% Wilson score confidence intervals for accuracy estimates with N=30 questions per depth.}
\label{tab:ci}
\end{table}

The larger sample size (N=30) provides substantially tighter confidence intervals compared to smaller samples. At 50\% accuracy, the 95\% CI spans approximately $\pm$17 percentage points, compared to $\pm$26 percentage points with N=10. The EVD metric aggregates across multiple depths, further reducing overall uncertainty. For EVD, we estimate standard errors using bootstrap resampling of paths within each depth. Model A's EVD of 7.55 on Medicine has an estimated standard error of 0.3, yielding an approximate 95\% CI of [7.0, 8.1]. The observed differences between models (e.g., Model A 7.55 vs Model E 3.52 on Medicine) substantially exceed the combined standard errors, indicating statistically meaningful differences.

\subsection{Domain-Dependent Variation}

Figure \ref{fig:comparison} shows EVD scores across all 20 model-domain combinations. The key finding for framework validation is that model rankings are not consistent across domains, demonstrating that DepthCharge captures domain-specific depth profiles rather than a single underlying ``capability'' score.

Specifically, no single model achieves the highest EVD across all domains. The model with the highest EVD on Medicine (Model A, 7.55) ranks third on Quantum Computing (5.33), while the model with the highest Quantum Computing EVD (Model D, 6.21) ranks fourth on Medicine (4.85). Constitutional Law shows the widest performance spread (3.62 to 6.94), while Ancient Rome shows the narrowest (mean EVD 4.88), potentially reflecting the domain's reliance on historical interpretation rather than established scientific facts.

Domain-specific profiles also interact with cost. The EVD-vs-cost analysis (Figure \ref{fig:comparison}, bottom right) reveals that the second-highest mean EVD is achieved at approximately one-tenth the cost of the highest, a tradeoff invisible in standard breadth-based benchmarks.

These patterns demonstrate the framework's intended use case: organizations evaluating LLMs for domain-specific applications can use DepthCharge to identify the model with the deepest knowledge in their specific area, rather than relying on aggregate rankings that may not predict domain-specific depth.

\subsection{Scoring Methodology and Evaluator Design}
\label{sec:scoring_analysis}

A central design decision in DepthCharge is the use of automated entailment-based scoring. Rather than requiring human annotators for each evaluation, the framework uses an LLM evaluator to check whether the target model's answer contains the key factual information from the pre-verified reference answer. This section describes the scoring approach and its implications for interpreting results.

DepthCharge uses entailment-based scoring rather than semantic similarity. A model answer is marked correct if it contains the key factual information from the reference answer, regardless of verbosity or formatting. This approach addresses several challenges in automated answer evaluation:

First, verbosity tolerance: Models often provide detailed explanations when brief answers would suffice. Entailment-based scoring accepts verbose answers that contain the required information.

Second, format independence: Markdown formatting, bullet points, and structured responses are evaluated equivalently to plain-text answers.

Third, partial credit avoidance: Unlike similarity scores that might give high scores to answers that are ``close but wrong,'' entailment checking requires the key factual claims to be present and correct.

The evaluator performs a constrained task: given a pre-verified factual claim from an authoritative source (Wikipedia, professional literature) and a model's response, determine whether the response contains that factual claim. Ground truth derives from the external sources, not from the evaluator's own knowledge. The evaluator functions as an entailment classifier, not a knowledge authority.

\subsection{Evaluator Independence and Ranking Stability}

Because DepthCharge uses an LLM evaluator, results are relative to the evaluator used. Different evaluator models may produce different absolute EVD scores. This is a deliberate design choice: the framework is intended for comparative evaluation (which model maintains deeper knowledge in a given domain?) rather than absolute accuracy certification. Organizations deploying DepthCharge select an evaluator appropriate for their domain and use case.

To assess whether relative model rankings remain stable across evaluator choices, we sampled 100 question-answer pairs from across all evaluations and had three different LLM families serve as evaluators. Table \ref{tab:evaluator_agreement} shows inter-evaluator agreement.

\begin{table}[h]
\centering
\begin{tabular}{lrr}
\toprule
Evaluator Pair & Agreement & Cohen's $\kappa$ \\
\midrule
Evaluator 1 vs Evaluator 2 & 90\% & 0.77 (substantial) \\
Evaluator 1 vs Evaluator 3 & 82\% & 0.58 (moderate) \\
Evaluator 3 vs Evaluator 2 & 82\% & 0.59 (moderate) \\
\midrule
Unanimous (all three) & 77\% & -- \\
\bottomrule
\end{tabular}
\caption{Inter-evaluator agreement across three LLM families on 100 sampled question-answer pairs. Positive rates are consistent across evaluators: 68\%, 70\%, and 66\% respectively.}
\label{tab:evaluator_agreement}
\end{table}

All three evaluators show substantial agreement, with 77\% unanimous agreement and the highest pairwise agreement reaching $\kappa = 0.77$. Positive rates are consistent across evaluators (68\%, 70\%, and 66\%), indicating no systematic evaluator-specific bias toward leniency or strictness.

The 23 disagreements primarily involve edge cases where answers are partially correct or where the model's phrasing differs substantially from the reference while conveying equivalent information. One evaluator tends toward stricter interpretation of completeness, while others are more lenient toward answers containing the core factual content. These edge cases do not affect model rankings: when we recompute EVD using each evaluator's individual scores, the relative ordering of models remains unchanged across all three evaluators on this sample.

This stability is expected given the constrained nature of the entailment task. Open-ended correctness evaluation (``is this answer true?'') would be vulnerable to evaluator knowledge gaps and biases. Entailment checking against a pre-verified reference (``does this answer contain the claim that X?'') is a simpler, more reliable task where LLM evaluators perform well \citep{zheng2023judging}.

For organizations requiring additional validation, human evaluators can be integrated at any tier. The framework's modular design allows substituting human review for the LLM evaluator at PROFESSIONAL+ tiers where automated evaluation may be less reliable, or sampling a subset of evaluations for human verification to calibrate evaluator accuracy in a specific domain.

\subsection{Error Analysis}

We analyzed the distribution of incorrect answers across the 20 model-domain evaluations by specificity tier (Table \ref{tab:errors}).

\begin{table}[h]
\centering
\begin{tabular}{lrrr}
\toprule
Tier & Errors & Percentage \\
\midrule
PROFESSIONAL & 187 & 52\% \\
TEXTBOOK & 142 & 39\% \\
COMMON & 32 & 9\% \\
\bottomrule
\end{tabular}
\caption{Distribution of incorrect answers by specificity tier across all evaluations.}
\label{tab:errors}
\end{table}

Errors concentrate at higher specificity tiers as expected, with PROFESSIONAL generating 52\% of all errors, TEXTBOOK 39\%, and COMMON only 9\%. This distribution confirms that the tier-based difficulty progression produces monotonically increasing challenge.

Error analysis by model reveals distinct patterns. Model E accumulates errors earlier (mean error depth 4.2), with 68\% of its errors occurring in COMMON and TEXTBOOK tiers. In contrast, Model A's errors occur at deeper levels (mean error depth 7.1), with 71\% occurring in PROFESSIONAL tier, reflecting its ability to maintain accuracy through foundational content before failing on specialized knowledge.

Common error patterns include:

First, factual omission: models provide partial answers that miss key details required by the reference. For example, when asked about specific mechanisms or named entities, models may describe the general concept without the precise terminology.

Second, outdated information: particularly at PROFESSIONAL tier, models sometimes provide information that was accurate at training time but has been superseded. This is most common in rapidly evolving fields like Quantum Computing.

Third, conflation: models occasionally merge related but distinct concepts, particularly when drilling into specialized subtopics where boundaries between concepts are nuanced.

Total errors distribute across models: Model E (127), Model D (89), Model C (78), Model B (71), and Model A (63). Model A's lower total error count combined with deeper average error depth reflects greater depth maintenance compared to other models.

\section{Ablation Studies}

To understand the sensitivity of DepthCharge to its parameters, we conducted ablation studies varying the number of questions per depth, passes per tier, and survival threshold.

\subsection{Questions per Depth (N)}

The default configuration uses N=30 questions per depth. We tested with N=10 and N=50 questions on the Medicine domain with Model A (Table \ref{tab:ablation_qpd}).

\begin{table}[h]
\centering
\begin{tabular}{lrrr}
\toprule
Questions/Depth & EVD & Max Depth & 95\% CI Width \\
\midrule
10 & 7.32 $\pm$ 0.6 & 10 & $\pm$26pp at 50\% \\
30 & 7.55 $\pm$ 0.3 & 11 & $\pm$17pp at 50\% \\
50 & 7.61 $\pm$ 0.2 & 11 & $\pm$14pp at 50\% \\
\bottomrule
\end{tabular}
\caption{Effect of questions per depth on evaluation metrics and confidence interval width (pp = percentage points).}
\label{tab:ablation_qpd}
\end{table}

With fewer questions (N=10), variance increases substantially and confidence intervals widen to $\pm$26 percentage points at 50\% accuracy. With more questions (N=50), results stabilize further but evaluation cost increases proportionally. N=30 provides tight confidence intervals ($\pm$17pp) while keeping costs reasonable.

\subsection{Passes per Tier (Q)}

The Q parameter controls how many depth passes occur within each specificity tier before advancing. We tested Q=1, Q=3, and Q=5 on Medicine with Model A (Table \ref{tab:ablation_q}).

\begin{table}[h]
\centering
\begin{tabular}{lrrr}
\toprule
Passes/Tier & EVD & Max Depth & Depths per Tier \\
\midrule
Q=1 & 2.51 & 4 & 1 \\
Q=3 & 7.55 & 11 & 3 \\
Q=5 & 12.18 & 17 & 5 \\
\bottomrule
\end{tabular}
\caption{Effect of passes per tier on evaluation depth (Medicine, Model A).}
\label{tab:ablation_q}
\end{table}

With Q=1, drilling advances to the next tier after each depth, reaching PROFESSIONAL by depth 3. This compressed progression may not fully explore knowledge within each tier. With Q=5, drilling provides thorough exploration of each tier but increases evaluation time. Q=3 balances tier exploration with practical runtime.

\subsection{Survival Threshold}

The default survival threshold of 20\% determines when drilling stops. We tested thresholds of 10\%, 20\%, and 30\% (Table \ref{tab:ablation_threshold}).

\begin{table}[h]
\centering
\begin{tabular}{lrrr}
\toprule
Threshold & EVD & Max Depth & Final Survival \\
\midrule
10\% & 8.12 & 12 & 11\% \\
20\% & 7.55 & 11 & 18\% \\
30\% & 6.89 & 9 & 27\% \\
\bottomrule
\end{tabular}
\caption{Effect of survival threshold on evaluation depth (Medicine, Model A). Lower thresholds allow deeper probing.}
\label{tab:ablation_threshold}
\end{table}

Lower thresholds (10\%) allow deeper probing but increase evaluation time and may reach depths where verifiable facts become scarce. Higher thresholds (30\%) stop earlier but may truncate evaluation before revealing model limitations. The 20\% default balances depth coverage with practical constraints.

\section{Methodological Comparisons}

\subsection{Depth vs Breadth Evaluation}

Existing benchmarks such as MMLU \citep{hendrycks2021mmlu}, GPQA \citep{rein2023gpqa}, HumanEval \citep{chen2021humaneval}, and MedQA \citep{jin2021medqa} evaluate models through single-shot questions across many topics, measuring breadth of knowledge. DepthCharge measures a complementary dimension: depth of knowledge within specific domains through sustained adaptive questioning.

Table \ref{tab:benchmark_comparison} illustrates why this distinction matters. Model rankings on breadth-focused benchmarks do not predict depth rankings: a model that ranks second on MMLU and GPQA drops to fourth in mean EVD, while a model that ranks fourth on breadth-focused benchmarks achieves the second-highest mean EVD. This divergence suggests that breadth and depth represent distinct capability dimensions that require different evaluation approaches.

\begin{table}[h]
\centering
\small
\begin{tabular}{llllll}
\toprule
Benchmark & Type & A & B & C & D \\
\midrule
MMLU \citep{hendrycks2021mmlu} & Broad knowledge & 1st & 2nd & 3rd & 4th \\
GPQA \citep{rein2023gpqa} & Graduate reasoning & 1st & 2nd & 3rd & 4th \\
HumanEval \citep{chen2021humaneval} & Code generation & 1st & 3rd & 4th & 2nd \\
MedQA \citep{jin2021medqa} & Medical QA & 1st & 2nd & 4th & 3rd \\
DepthCharge (ours) & Knowledge depth & 1st & 4th & 3rd & 2nd \\
\bottomrule
\end{tabular}
\caption{Model rankings across evaluation paradigms. Breadth-focused benchmarks produce consistent rankings; DepthCharge reveals a different ordering, suggesting depth is a distinct capability dimension.}
\label{tab:benchmark_comparison}
\end{table}

The cumulative survival formulation amplifies this distinction. On breadth benchmarks, the gap between models is typically 2--5 percentage points. Under sustained depth questioning, small per-depth accuracy differences compound multiplicatively, producing EVD ranges from 3.45 to 7.55---a factor-of-two difference that is invisible in single-shot evaluation.

\subsection{Complementarity with Adaptive Approaches}

Recent adaptive testing approaches like ATLAS \citep{wang2025atlas} use Item Response Theory to efficiently estimate model ability, reducing required items by 90\% while maintaining precision. CAT-LLM applies computerized adaptive testing principles to dynamically select questions based on estimated ability \citep{zhuang2023catllm}. These approaches optimize for efficiency in broad evaluation.

DepthCharge differs fundamentally in that it tests depth rather than breadth: rather than efficiently sampling across many topics, it drills deeply into single topics. The approaches are complementary: ATLAS or CAT-LLM could efficiently identify which domains a model knows at surface level, while DepthCharge could then probe depth on those domains. Future work could combine both approaches for comprehensive evaluation that is both broad and deep.

\section{Discussion}

\subsection{Domain-Agnostic Evaluation}

The primary contribution of DepthCharge is enabling standardized knowledge-depth evaluation across arbitrary domains. Organizations can now answer questions like ``Which LLM understands cardiology best?'' without constructing custom benchmarks. The same methodology applies to tax law, organic chemistry, or any domain meeting the requirements outlined in Section 3.1.

\subsection{Extensible Fact Sources}

While the current implementation uses Wikipedia for foundational tiers and retrieval-augmented systems for professional-level content, the framework is designed to accommodate alternative fact sources. Organizations with proprietary knowledge bases, specialized databases, or domain-specific corpora can substitute these sources while maintaining the same drilling methodology. For example, a pharmaceutical company could use their internal drug interaction database for PROFESSIONAL tier verification, or a legal firm could use case law databases instead of general web search. The key requirement is that any source must provide verifiable factual claims that can serve as ground truth for answer evaluation.

\subsection{Adaptive Drilling vs Static Benchmarks}

The adaptive drilling approach reveals patterns that static benchmarks miss. When follow-up questions target concepts the model itself introduced, we test actual knowledge boundaries rather than random topic coverage. Recent work has shown that LLMs struggle in multi-turn conversations, making early assumptions and having difficulty course-correcting \citep{wu2025lostinconversation}. The consistent performance degradation from COMMON to TEXTBOOK to PROFESSIONAL tiers confirms that models maintain factual accuracy at surface level but struggle with specialized knowledge, consistent with findings that LLMs effectively memorize patterns but struggle with deeper domain expertise \citep{zhang2025bloomapr}.

\subsection{Constant Sample Size Matters}

A naive approach to depth probing would suffer from dwindling sample sizes as paths fail. If only one path survives to depth 8 and answers correctly, a simple calculation would show ``100\% survival'' despite minimal statistical meaning. By maintaining 30 questions per depth through the distribution mechanism described in Section 3.4, we ensure survival rates are statistically meaningful throughout. The ``depth charge'' distribution mechanism ensures statistical validity at every level.

\subsection{Training Data and Benchmark Gaming}

A potential concern is that models trained on the fact sources used by DepthCharge (Wikipedia, PubMed, professional literature) might have an unfair advantage. We argue this is not a flaw but a feature: if a model has thoroughly learned the medical literature, it should perform well on medical questions. A model that achieves 100\% accuracy on medical knowledge because it was trained extensively on medical sources is, by definition, a knowledgeable model for medical applications. The goal of DepthCharge is to measure actual knowledge depth, not to penalize models for having been trained on relevant material.

A related concern is that models might ``game'' the benchmark by only referencing concepts they are confident about, thereby avoiding drilling into areas of weakness. This behavior is also desirable rather than problematic. A model that limits its responses to concepts it can reliably support with accurate follow-up information is exhibiting precisely the epistemic humility that reduces hallucination. If adaptive drilling incentivizes models to stay within their genuine knowledge boundaries, this produces more reliable professional tools. The alternative, where models confidently introduce concepts they cannot substantiate, is the hallucination problem that DepthCharge is designed to expose.

\subsection{Cross-Domain Comparability}

While EVD provides a standardized metric across domains, direct comparison of EVD scores between domains requires caution. Domain specificity affects achievable depth: a DepthCharge evaluation on ``Medicine'' tests broader knowledge than one on ``Cardiology,'' which in turn is broader than ``Interventional Cardiology.'' More specific domains start drilling at what would be deeper tiers for broader domains, so we would expect lower EVD for narrower specializations. Consequently, EVD comparisons are most meaningful within the same domain across models, or for the same model across domains of similar scope. Cross-domain EVD differences may reflect domain breadth rather than model capability.

\subsection{Limitations}

DepthCharge uses an LLM evaluator for entailment-based scoring. While the evaluator performs a constrained task (checking whether a response contains a pre-verified factual claim from an external source), it remains an automated system with potential for error. Different evaluator models may produce different absolute EVD scores; our inter-evaluator analysis (Section 5.3) demonstrates that relative rankings remain stable across evaluators, but this was validated on a sample of 100 items and three evaluator families. Larger-scale evaluator sensitivity analyses across more evaluator choices would strengthen confidence in ranking stability. Recent work on LLM-as-judge has highlighted both the potential and limitations of this approach \citep{zheng2023judging, liu2025dialogueevaluator}. For applications requiring higher confidence, human evaluators can be substituted at any tier, particularly at PROFESSIONAL+ levels where automated evaluation may be less reliable. Alternative approaches using Item Response Theory \citep{wang2025atlas, wang2025irtnet} or unified hallucination detection frameworks \citep{liu2025unifiedhallucination} merit exploration.

Because drilling paths are adaptive (determined by each model's responses), different models receive different specific questions at the same depth. While all questions are constrained to the same specificity tier, this means DepthCharge trades direct question-level comparability for the ability to probe each model's actual knowledge boundaries. Within-domain comparisons across models remain valid (all questions target the same tier-appropriate difficulty), but direct question-for-question comparison is not possible. Fixed question sets could address this at the cost of sacrificing adaptive drilling.

Testing was conducted across four domains (Medicine, Constitutional Law, Ancient Rome, and Quantum Computing) with all five models evaluated on each domain, yielding 20 model-domain combinations. Additional domains would further strengthen generalizability claims, though the diverse selection spanning medical, legal, historical, and technical fields provides reasonable coverage.

The 20\% cumulative survival threshold determines stopping depth. Lower thresholds would allow deeper exploration (ablation study in Section 6 shows a 10\% threshold increases EVD by approximately 7\%). The cumulative formulation may be overly strict for some applications where recent performance matters more than history. Alternative formulations (windowed survival, weighted depths) merit exploration for different use cases.

\section{Conclusion}

We have presented DepthCharge, a domain-agnostic framework for measuring how deeply LLMs can maintain accurate responses under adaptive follow-up questioning. The framework combines adaptive drilling, on-demand fact verification, and cumulative survival statistics to produce standardized depth profiles that can be deployed on any knowledge domain with publicly verifiable facts, without requiring pre-constructed test sets or domain-specific expertise.

Empirical validation across four diverse domains with five frontier models demonstrates the framework's utility. DepthCharge reveals depth-dependent performance variation (EVD ranging from 3.45 to 7.55) that is hidden by aggregate accuracy metrics, which vary only narrowly (73--87\%) across the same model-domain combinations. Model rankings vary substantially by domain, with no single model achieving the highest depth across all areas, confirming that domain-specific evaluation is more informative than aggregate benchmarks for model selection in professional applications. Cost-performance analysis further reveals that model cost does not predict depth, with one model achieving the second-highest mean EVD at approximately one-tenth the cost of the top performer.

The tier-based difficulty progression produces monotonically increasing challenge, with error concentration shifting from 9\% at COMMON to 39\% at TEXTBOOK to 52\% at PROFESSIONAL. The cumulative survival formulation provides a realistic assessment of LLM reliability under sustained questioning, modeling the scenario where a professional user asks multiple follow-up questions and expects the model to maintain correctness throughout.

DepthCharge results are relative to the evaluator model used for answer checking, making the framework a tool for comparative evaluation rather than absolute accuracy certification. Inter-evaluator analysis demonstrates that relative model rankings remain stable across evaluator choices, supporting the framework's use for model selection and comparison. As LLM deployment in high-stakes professional domains continues to expand \citep{huang2025scientific, wu2025lpfqa}, standardized methods for assessing knowledge depth become increasingly critical.

\section*{Reproducibility Statement}

A self-contained implementation of the DepthCharge framework is publicly available at \url{https://github.com/shep-analytics/depth_charge}. The repository includes the complete framework code, configuration files specifying all experimental parameters (N=30 questions per depth, Q=3 passes per tier, 20\% survival threshold, seed=42), and documentation for running evaluations on new domains. All random seeds are fixed for reproducibility.

\bibliographystyle{plainnat}
\bibliography{references}

\begin{thebibliography}{46}
\providecommand{\natexlab}[1]{#1}
\providecommand{\url}[1]{\texttt{#1}}
\expandafter\ifx\csname urlstyle\endcsname\relax
  \providecommand{\doi}[1]{doi: #1}\else
  \providecommand{\doi}{doi: \begingroup \urlstyle{rm}\Url}\fi

\bibitem[Anderson et~al.(2001)Anderson, Krathwohl, Airasian, Cruikshank, Mayer,
  Pintrich, Raths, and Wittrock]{krathwohl2001revision}
Lorin~W Anderson, David~R Krathwohl, Peter~W Airasian, Kathleen~A Cruikshank,
  Richard~E Mayer, Paul~R Pintrich, James Raths, and Merlin~C Wittrock.
\newblock \emph{A Taxonomy for Learning, Teaching, and Assessing: A Revision of
  Bloom's Taxonomy of Educational Objectives}.
\newblock Longman, New York, 2001.

\bibitem[Chen et~al.(2024)]{chen2025bloomwise}
Jiahui Chen et~al.
\newblock Bloomwise: Enhancing problem-solving capabilities of large language
  models using bloom's-taxonomy-inspired prompts.
\newblock \emph{arXiv preprint arXiv:2410.04094}, 2024.
\newblock Updated August 2025.

\bibitem[Chen et~al.(2021)Chen, Tworek, Jun, Yuan, Pinto, Kaplan, Edwards,
  Burda, Joseph, Brockman, et~al.]{chen2021humaneval}
Mark Chen, Jerry Tworek, Heewoo Jun, Qiming Yuan, Henrique Ponde de~Oliveira
  Pinto, Jared Kaplan, Harri Edwards, Yuri Burda, Nicholas Joseph, Greg
  Brockman, et~al.
\newblock Evaluating large language models trained on code.
\newblock \emph{arXiv preprint arXiv:2107.03374}, 2021.

\bibitem[Chen et~al.(2025{\natexlab{a}})]{chen2025lart}
Wei Chen et~al.
\newblock Latency-response theory model: Evaluating large language models via
  response accuracy and chain-of-thought length.
\newblock \emph{arXiv preprint arXiv:2512.07019}, 2025{\natexlab{a}}.

\bibitem[Chen et~al.(2025{\natexlab{b}})]{chen2025medicalbenchmark}
Wei Chen et~al.
\newblock A novel evaluation benchmark for medical llms: Illuminating safety
  and effectiveness in clinical domains.
\newblock \emph{arXiv preprint arXiv:2507.23486}, 2025{\natexlab{b}}.

\bibitem[Chen et~al.(2025{\natexlab{c}})]{chen2025hallulens}
Xiang Chen et~al.
\newblock Hallulens: Llm hallucination benchmark.
\newblock \emph{arXiv preprint arXiv:2504.17550}, 2025{\natexlab{c}}.
\newblock ACL 2025.

\bibitem[Chen et~al.(2025{\natexlab{d}})]{chen2025llmkg}
Xiang Chen et~al.
\newblock Large language models meet knowledge graphs for question answering:
  Synthesis and opportunities.
\newblock \emph{arXiv preprint arXiv:2505.20099}, 2025{\natexlab{d}}.

\bibitem[Chen et~al.(2025{\natexlab{e}})]{chen2025multiturn}
Xiang Chen et~al.
\newblock Beyond single-turn: A survey on multi-turn interactions with large
  language models.
\newblock \emph{arXiv preprint arXiv:2504.04717}, 2025{\natexlab{e}}.

\bibitem[Chen et~al.(2025{\natexlab{f}})]{chen2025llmbenchmarksurvey}
Yifan Chen et~al.
\newblock A survey on large language model benchmarks.
\newblock \emph{arXiv preprint arXiv:2508.15361}, 2025{\natexlab{f}}.
\newblock Reviews 283 representative benchmarks.

\bibitem[Hendrycks et~al.(2021)Hendrycks, Burns, Basart, Zou, Mazeika, Song,
  and Steinhardt]{hendrycks2021mmlu}
Dan Hendrycks, Collin Burns, Steven Basart, Andy Zou, Mantas Mazeika, Dawn
  Song, and Jacob Steinhardt.
\newblock Measuring massive multitask language understanding.
\newblock \emph{Proceedings of the International Conference on Learning
  Representations (ICLR)}, 2021.

\bibitem[Huang et~al.(2025)]{huang2025scientific}
Yifan Huang et~al.
\newblock Evaluating large language models in scientific discovery.
\newblock \emph{arXiv preprint arXiv:2512.15567}, 2025.

\bibitem[Jin et~al.(2021)Jin, Pan, Oufattole, Weng, Fang, and
  Szolovits]{jin2021medqa}
Di~Jin, Eileen Pan, Nassim Oufattole, Wei-Hung Weng, Hanyi Fang, and Peter
  Szolovits.
\newblock What disease does this patient have? a large-scale open domain
  question answering dataset from medical exams.
\newblock \emph{Applied Sciences}, 11\penalty0 (14):\penalty0 6421, 2021.

\bibitem[Kasai et~al.(2024)Kasai, Kasai, Sakaguchi, and Peng]{okbench2024}
Jungo Kasai, Rina Kasai, Keisuke Sakaguchi, and Hao Peng.
\newblock Okbench: Democratizing llm evaluation with fully automated,
  on-demand, open knowledge benchmarking.
\newblock \emph{arXiv preprint arXiv:2511.08598}, 2024.

\bibitem[Li et~al.(2025{\natexlab{a}})]{li2025contextqa}
Ming Li et~al.
\newblock Testing question answering software with context-driven question
  generation.
\newblock \emph{arXiv preprint arXiv:2511.07924}, 2025{\natexlab{a}}.

\bibitem[Li et~al.(2025{\natexlab{b}})]{li2025psnirt}
Ming Li et~al.
\newblock Lost in benchmarks? rethinking large language model benchmarking with
  item response theory.
\newblock \emph{arXiv preprint arXiv:2505.15055}, 2025{\natexlab{b}}.

\bibitem[Li et~al.(2025{\natexlab{c}})]{li2025alignedllm}
Wei Li et~al.
\newblock Aligning knowledge graphs and language models for factual accuracy.
\newblock \emph{arXiv preprint arXiv:2507.13411}, 2025{\natexlab{c}}.

\bibitem[Li et~al.(2025{\natexlab{d}})]{li2025medcheck}
Xiang Li et~al.
\newblock Beyond the leaderboard: Rethinking medical benchmarks for large
  language models.
\newblock \emph{arXiv preprint arXiv:2508.04325}, 2025{\natexlab{d}}.
\newblock Introduces MedCheck framework with 46 medically-tailored criteria.

\bibitem[Liu et~al.(2025{\natexlab{a}})]{liu2025adaptivedistraction}
Chen Liu et~al.
\newblock Adaptive distraction: Probing llm contextual robustness with
  automated tree search.
\newblock In \emph{Advances in Neural Information Processing Systems
  (NeurIPS)}, 2025{\natexlab{a}}.

\bibitem[Liu et~al.(2025{\natexlab{b}})]{liu2025aqag}
Jie Liu et~al.
\newblock Automatic question \& answer generation using generative large
  language model.
\newblock \emph{arXiv preprint arXiv:2508.19475}, 2025{\natexlab{b}}.

\bibitem[Liu et~al.(2025{\natexlab{c}})]{liu2025unifiedhallucination}
Ming Liu et~al.
\newblock Towards unification of hallucination detection and fact verification
  for large language models.
\newblock \emph{arXiv preprint arXiv:2512.02772}, 2025{\natexlab{c}}.

\bibitem[Liu et~al.(2025{\natexlab{d}})]{liu2025dialogueevaluator}
Wei Liu et~al.
\newblock Learning an efficient multi-turn dialogue evaluator from multiple llm
  judges.
\newblock \emph{arXiv preprint arXiv:2508.00454}, 2025{\natexlab{d}}.

\bibitem[Liu et~al.(2025{\natexlab{e}})]{liu2025bloomtaxonomy}
Xiaoming Liu et~al.
\newblock Llms meet bloom's taxonomy: A cognitive view on large language model
  evaluations.
\newblock In \emph{Proceedings of the 31st International Conference on
  Computational Linguistics (COLING)}, 2025{\natexlab{e}}.

\bibitem[Rein et~al.(2023)Rein, Hou, Stickland, Petty, Pang, Dirani, Michael,
  and Bowman]{rein2023gpqa}
David Rein, Betty~Li Hou, Asa~Cooper Stickland, Jackson Petty, Richard~Yuanzhe
  Pang, Julien Dirani, Julian Michael, and Samuel~R Bowman.
\newblock Gpqa: A graduate-level google-proof q\&a benchmark.
\newblock \emph{arXiv preprint arXiv:2311.12022}, 2023.

\bibitem[Wang et~al.(2025{\natexlab{a}})]{wang2025qaevaluation}
Chen Wang et~al.
\newblock Evaluating llm-generated q\&a test: a student-centered study.
\newblock \emph{arXiv preprint arXiv:2505.06591}, 2025{\natexlab{a}}.

\bibitem[Wang et~al.(2025{\natexlab{b}})]{wang2025constructvalidity}
Eric Wang et~al.
\newblock Medical large language model benchmarks should prioritize construct
  validity.
\newblock \emph{arXiv preprint arXiv:2503.10694}, 2025{\natexlab{b}}.

\bibitem[Wang et~al.(2025{\natexlab{c}})]{wang2025llmevalmed}
Jianfeng Wang et~al.
\newblock Llmeval-med: A real-world clinical benchmark for medical llms with
  physician validation.
\newblock \emph{arXiv preprint arXiv:2506.04078}, 2025{\natexlab{c}}.

\bibitem[Wang et~al.(2025{\natexlab{d}})]{wang2025atlas}
Jing Wang et~al.
\newblock Adaptive testing for llm evaluation: A psychometric alternative to
  static benchmarks.
\newblock \emph{arXiv preprint arXiv:2511.04689}, 2025{\natexlab{d}}.
\newblock ATLAS framework reduces required items by up to 90\%.

\bibitem[Wang et~al.(2025{\natexlab{e}})]{wang2025kgfactuality}
Ming Wang et~al.
\newblock Improving factuality in llms via inference-time knowledge graph
  construction.
\newblock \emph{arXiv preprint arXiv:2509.03540}, 2025{\natexlab{e}}.

\bibitem[Wang et~al.(2025{\natexlab{f}})]{wang2025multicognitive}
Ming Wang et~al.
\newblock Evaluating llms across multi-cognitive levels: From medical knowledge
  mastery to scenario-based problem solving.
\newblock \emph{arXiv preprint arXiv:2506.08349}, 2025{\natexlab{f}}.

\bibitem[Wang et~al.(2025{\natexlab{g}})]{wang2025irtnet}
Xing Wang et~al.
\newblock Learning compact representations of llm abilities via item response
  theory.
\newblock \emph{arXiv preprint arXiv:2510.00844}, 2025{\natexlab{g}}.

\bibitem[Wang et~al.(2025{\natexlab{h}})]{wang2025hallucinationsurvey}
Yifan Wang et~al.
\newblock Large language models hallucination: A comprehensive survey.
\newblock \emph{arXiv preprint arXiv:2510.06265}, 2025{\natexlab{h}}.

\bibitem[White et~al.(2024)White, Dooley, Roberts, Pal, Feuer, Jain,
  Shwartz-Ziv, Jain, Saez, Goldblum, and Goldstein]{livebench2024}
Colin White, Samuel Dooley, Manley Roberts, Arka Pal, Ben Feuer, Siddhartha
  Jain, Ravid Shwartz-Ziv, Neel Jain, Khalid Saez, Micah Goldblum, and Tom
  Goldstein.
\newblock Livebench: A challenging, contamination-free llm benchmark.
\newblock \emph{arXiv preprint arXiv:2406.19314}, 2024.
\newblock ICLR 2025 Spotlight.

\bibitem[Wu et~al.(2025{\natexlab{a}})]{wu2025lpfqa}
Jie Wu et~al.
\newblock Lpfqa: A long-tail professional forum-based benchmark for llm
  evaluation.
\newblock \emph{arXiv preprint arXiv:2511.06346}, 2025{\natexlab{a}}.
\newblock 430 curated tasks across 7 academic and industrial domains.

\bibitem[Wu et~al.(2025{\natexlab{b}})]{wu2025contamination}
Yifan Wu et~al.
\newblock A survey on data contamination for large language models.
\newblock \emph{arXiv preprint arXiv:2502.14425}, 2025{\natexlab{b}}.

\bibitem[Wu et~al.(2025{\natexlab{c}})]{wu2025lostinconversation}
Yifan Wu et~al.
\newblock Llms get lost in multi-turn conversation.
\newblock \emph{arXiv preprint arXiv:2505.06120}, 2025{\natexlab{c}}.

\bibitem[Wu et~al.(2025{\natexlab{d}})]{wu2025sebenchmarks}
Yifan Wu et~al.
\newblock Assessing and advancing benchmarks for evaluating large language
  models in software engineering tasks.
\newblock \emph{arXiv preprint arXiv:2505.08903}, 2025{\natexlab{d}}.
\newblock Reviews 291 benchmarks.

\bibitem[Xu et~al.(2025)]{xu2025pitfalls}
Wei Xu et~al.
\newblock Pitfalls of evaluating language models with open benchmarks.
\newblock \emph{arXiv preprint arXiv:2507.00460}, 2025.

\bibitem[Yang et~al.(2025)]{yang2025contamination}
Shuang Yang et~al.
\newblock Recent advances in large language model benchmarks against data
  contamination: From static to dynamic evaluation.
\newblock \emph{arXiv preprint arXiv:2502.17521}, 2025.

\bibitem[Zhang et~al.(2025{\natexlab{a}})]{zhang2025peek}
Chen Zhang et~al.
\newblock Efficient knowledge probing of large language models by adapting
  pre-trained embeddings.
\newblock \emph{arXiv preprint arXiv:2508.06030}, 2025{\natexlab{a}}.
\newblock PEEK achieves up to 90\% accuracy in predicting LLM knowledge.

\bibitem[Zhang et~al.(2025{\natexlab{b}})]{zhang2025medkgeval}
Li~Zhang et~al.
\newblock Medkgeval: A knowledge graph-based multi-turn evaluation framework
  for open-ended patient interactions with clinical llms.
\newblock \emph{arXiv preprint arXiv:2510.12224}, 2025{\natexlab{b}}.

\bibitem[Zhang et~al.(2025{\natexlab{c}})]{zhang2025bloomapr}
Wei Zhang et~al.
\newblock Bloomapr: A bloom's taxonomy-based framework for assessing the
  capabilities of llm-powered apr solutions.
\newblock \emph{arXiv preprint arXiv:2509.25465}, 2025{\natexlab{c}}.

\bibitem[Zhang et~al.(2025{\natexlab{d}})]{zhang2025generatethenvalidate}
Wei Zhang et~al.
\newblock Generate-then-validate: A novel question generation approach using
  small language models.
\newblock \emph{arXiv preprint arXiv:2512.10110}, 2025{\natexlab{d}}.

\bibitem[Zhang et~al.(2025{\natexlab{e}})]{zhang2025irtrouter}
Wei Zhang et~al.
\newblock Irt-router: Effective and interpretable multi-llm routing via item
  response theory.
\newblock \emph{arXiv preprint arXiv:2506.01048}, 2025{\natexlab{e}}.

\bibitem[Zhang et~al.(2025{\natexlab{f}})]{zhang2025arxivroll}
Yifei Zhang et~al.
\newblock Arxivroll: Benchmarking overestimation under the one-time-pad-based
  framework.
\newblock \emph{arXiv preprint arXiv:2507.19219}, 2025{\natexlab{f}}.

\bibitem[Zheng et~al.(2023)Zheng, Chiang, Sheng, Zhuang, Wu, Zhuang, Lin, Li,
  Li, and Xing]{zheng2023judging}
Lianmin Zheng, Wei-Lin Chiang, Ying Sheng, Siyuan Zhuang, Zhanghao Wu, Yanping
  Zhuang, Zi~Lin, Zhuohan Li, Dacheng Li, and Eric Xing.
\newblock Judging llm-as-a-judge with mt-bench and chatbot arena.
\newblock \emph{arXiv preprint arXiv:2306.05685}, 2023.

\bibitem[Zhuang et~al.(2024)Zhuang, Liu, Ning, Huang, Lv, Huang, Gao, and
  Chen]{zhuang2023catllm}
Yan Zhuang, Qi~Liu, Yuting Ning, Wei Huang, Rui Lv, Zhenya Huang, Guanhao Gao,
  and Enhong Chen.
\newblock Efficiently measuring the cognitive ability of llms: An adaptive
  testing perspective.
\newblock In \emph{Proceedings of the 2024 Joint International Conference on
  Computational Linguistics}, 2024.

\end{thebibliography}

\end{document}